\documentclass[sn-mathphys,Numbered]{sn-jnl}

\usepackage{graphicx}%
\usepackage{multirow}%
\usepackage{amsmath,amssymb,amsfonts}%
\usepackage{amsthm}%
\usepackage{mathrsfs}%
\usepackage[title]{appendix}%
\usepackage{xcolor}%
\usepackage{textcomp}%
\usepackage{manyfoot}%
\usepackage{booktabs}%
\usepackage{algorithm}%
\usepackage{algorithmicx}%
\usepackage{algpseudocode}%
\usepackage{listings}%
\usepackage{CJKutf8}
\usepackage{subcaption}
\usepackage{lipsum}

\newcommand{\predictiveImplication}{\mathrel{\Rightarrow\!\!\!\!\!\!\!/ \ }}
\newcommand{\retrospectiveImplication}{\mathrel{\Rightarrow\!\!\!\!\!\!\!\backslash \ }}

\newcommand{\predictiveEquivalance}{\mathrel{\Leftrightarrow\!\!\!\!\!\!/ \ }}

\theoremstyle{thmstyleone}%
\theoremstyle{thmstyletwo}%

\theoremstyle{thmstylethree}%

\begin{document}

\title[[A Brain-Inspired Sequence Learning Model based on a Logic]{A Brain-Inspired Sequence Learning Model\\based on a Logic}


\author{\fnm{Bowen} \sur{Xu}}\email{bowen.xu@temple.edu}


\affil{\orgdiv{Department of Computer and Information}, \orgname{Temple University}, \orgaddress{\street{1801 N Broad St}, \city{Philadelphia}, \postcode{19122}, \state{PA}, \country{USA}}}




\abstract{
Sequence learning is an essential aspect of intelligence. In Artificial Intelligence, sequence \textit{prediction} task is usually used to test a sequence learning model.
In this paper, a model of sequence learning, which is interpretable through Non-Axiomatic Logic, is designed and tested. The learning mechanism is composed of three steps, hypothesizing, revising, and recycling, which enable the model to work under the \textit{Assumption of Insufficient Knowledge and Resources}.
Synthetic datasets for sequence \textit{prediction} task are generated to test the capacity of the model. The results show that the model works well within different levels of difficulty. In addition, since the model adopts \textit{concept-centered} representation, it theoretically does not suffer from \textit{catastrophic forgetting}, and the practical results also support this property.
This paper shows the potential of learning sequences in a logical way.
}

\maketitle

\section{Introduction}\label{sec_inroduction}

\textit{Sequence leaning} (sometimes known as  \textit{sequential learning}, \textit{serial order learning}, \textit{etc.}) refers to acquiring the proper ordering of \textit{events} or stimuli~\cite{conway2012seq, sun2001sequence}. It is the foundation of many learning processes for an intelligent agent to interact with the world, such as sensorimotor process, natural language acquisition, \textit{etc}. 

In Cognitive Science, \textit{Serial Reaction-Time} task was widely used for measuring subjects' performance of sequence learning~\cite{clegg1998sequence}, where given some repeated sequences of stimuli, subjects' reaction time decreases with time goes by. While in Artificial Intelligence (AI), people usually measure the anticipation accuracy of a sequence learning model. There are several types of tasks in AI to evaluate a sequence learning model, including sequence \textit{prediction}, \textit{generation}, \textit{recognition}, and \textit{decision making}~\cite{sun2001sequence}. Various of approaches for sequence learning are proposed, including Markovian approaches~\cite{rabiner1986hmm}, recurrent neural network~\cite{medsker2001rnn}, \textit{etc}. Neural networks, such as Transformer~\cite{vaswani2017transformer}, have gained huge progress in natural language processing, which could be viewed as a special case of sequence learning task. There are some biologically plausible models, among which an intriguing one is Hierarchical Temporal Memory (HTM): through modeling neocortical column, the HTM model can memorize frequently occurring sequences as long as each \textit{event} can be converted to Sparse Distributed Representation (SDR) \cite{hawkins2016seq-mem}, though how to deal with uncertainty is still a challenge for the HTM model.

Explainability is an important issue on AI security, and a major criticism of neural networks is their lack of explainability: the models are black or grey boxes, and developers are hard to understand what is going on and how to fix it when unexpected behaviors occur. It can be argued that this issue can be addressed if a model follows a logic, in other words, a model is interpretable if it is described through symbolic or \textit{logical representation}. Among various candidates besides the well-known First-Order Predicated Logic (FOPL) and Expert Systems, there is a promising model of intelligent reasoning, Non-Axiomatic Logic (NAL)~\cite{wang2013nal}, which is able to deal with uncertainty and has proposed a solution of the \textit{symbol grounding} problem~\cite{harnad1990symbol_grounding, wang2005semantics}. 
In NAL, there are some logical rules for temporal inference\cite{wang2015temporal}, \textit{e.g.}, \textit{deduction}, \textit{induction}, \textit{etc}. However, how to extract temporal patterns from sequences remains a hard problem with this logical representation.


Highly inspired by HTM, the biologically-constrained model, and NAL, the logic for modeling intelligence, in this paper, a model of sequence learning is proposed. In HTM, a collection of mini-columns represents an event, and a neuron in a mini-column corresponds to a certain context. With the same intuition, the model proposed here has the structure of \textit{mini-column}~\cite{buxhoeveden2002minicolumn} but adopts \textit{concept-centered representation} (see Sec.~\ref{sec:logic-repr}) instead of distributed representation: each \textit{event} corresponds to a single \textit{mini-column}. In this paper, the model can be interpreted by NAL: a link between two neurons is interpreted as a statement of temporal implication/equivalence with a \textit{truth-value}. The strength of a link is modified via \textit{temporal induction}, and future events are anticipated by \textit{temporal deduction}. A \textit{mini-column} corresponds to a \textit{concept} in NAL, and a neuron's being activated corresponds to partial \textit{meanings} of the \textit{concept} being recalled. Due to the properties of NAL~\cite{wang2001uncertainty, wang2013nal}, the model is naturally capable of handling uncertainty, and the model's behaviors and internals are fully understandable by human beings.

The model is tested on \textit{prediction} tasks, where the input is a list of events, and the model is expected to predict future events. The list is assumed to have no beginning and no end (though in practice, usually there has to be a start-point), so that it is impossible for the model to memorize all the contents. With this assumption, the learning procedure should be \textit{online} and \textit{life-long}~\cite{hoi2021online}. 
An example of input is ``$(...,\$, A,B,C,D, \$,\$,X,B,C,Y,\$,...)$'', where ``$\$$'' denotes a random \textit{event}, while characters denote different types of events. It is noted that the types of \textit{events} are not predetermined before a system is initialized but dynamically constructed by the model. In this example, there are two prototypes of sequences, ``$(A,B,C,D)$'' and ``$(X,B,C,Y)$'', meaning that sequence ``$(A,B,C)$'' is always followed by event $D$, but by observing only ``$(B,C)$'' either $D$ or $Y$ is probable to occur immediately. In Sec.~\ref{sec_results}, the lengths and the number of prototypes vary in several cases, in order to test the capacity of the model. In the meanwhile, \textit{catastrophic forgetting}~\cite{mccloskey1989cf} is a difficult problem in models with distributed representation (\textit{e.g.}, in neural networks). The qualitative results show that the model proposed does not suffer from \textit{catastrophic forgetting}.

\section{Methods}\label{sec_methods}

The model is highly inspired by the \textit{mini-column} structure in neocortex~\cite{buxhoeveden2002minicolumn} as well as \textit{Non-Axiomatic Logic} (NAL)~\cite{wang2013nal}, a logic which can handle uncertainty. A \textit{mini-column} is viewed as a \textit{concept} in NAL, while a neuron inside the \textit{mini-column} is the same concept but with a special meaning under a certain context. A concept under different contexts has quite distinct meanings. For example, consider the following two sentences: 1) ``I go to the bank every month to save money.'' 2) ``This restaurant is located on the bank of the river.'' The word ``bank'' has different meanings under these two contexts. We can either say the same word corresponds to two different concepts, or say the same concept has different meanings. The two statements have no difference in practice. If there is no context, all the neurons in a \textit{mini-column} would be activated, in other words, all the meanings of the concept are taken into consideration. However, if there is a context, one neuron in the \textit{mini-column} would be \textit{pre-activated} and then \textit{activated}, in other words, the concept with a special meaning is anticipated and then activated. We can see there is a very natural correspondence between the \textit{mini-column} structure and \textit{concept}. For ease of description, a \textit{concept} with a special meaning under a certain context is referred to by the term \textit{contextual-concept}.

Based on this view, to learn a sequence is to connect a collection of \textit{contextual-concepts} one by one, while \textit{sequence learning} is on the learning algorithm to construct representations of sequences, \textit{i.e.}, \textit{chains} of \textit{conceptual-concepts}, given a list of \textit{events}. The list here has no explicit head and tail, \textit{i.e.}, it is endless. Consequently, the learning algorithm should not be \textit{offline} but \textit{online}, meaning that it is impossible for a machine accurately know all events of future and past. This is an actual situation humans meet in daily life. In this paper, a model is designed to deal with this situation. The model is brain-inspired, so that it can be described as a brain-like structure; in the mean while, more crucially, the model is based on a logic, so that it can also be interpreted into human-understandable knowledge. 
The design of such a model is described in details in the following.
\begin{figure}
    \centering
    \includegraphics[width=0.8\linewidth]{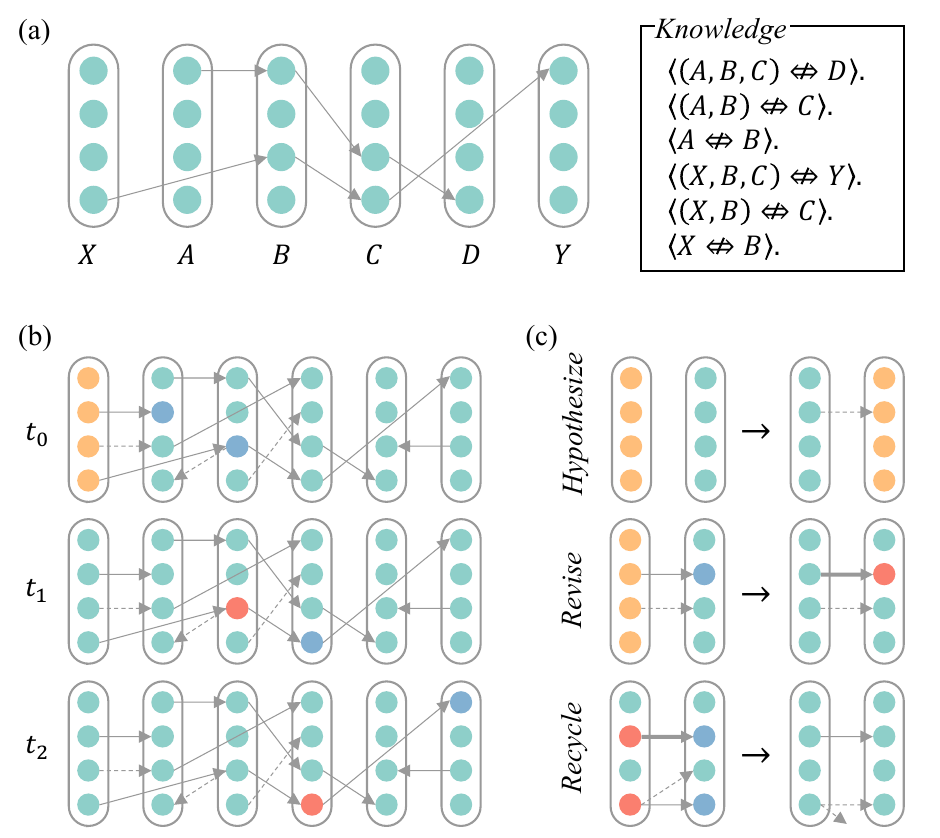}
    \caption{Model diagram. 
    (a) An example of the learned network. There are six \textit{concepts}, $A$ through $D$, $X$, and $Y$. Each \textit{concept} is represented as a \textit{column} that contains multiple \textit{nodes}. There are \textit{links} between \textit{nodes}. Multiple \textit{links} constitute a \textit{chain}, representing a group of knowledge. For instance,  \textit{chain} ``$(A^{(1)}, B^{(1)}, C^{(3)}, D^{(4)})$'' represents three beliefs, ``$\langle (A, B, C) \predictiveEquivalance D \rangle.$'', ``$\langle (A, B) \predictiveEquivalance C \rangle.$'', and ``$\langle A \predictiveEquivalance B \rangle.$''. 
    (b) An example of inference procedure. It shows the internals of the model at three consecutive time-steps. At $t_0$, \textit{concept} $X$ is activated. Since there is no context, all the \textit{nodes} are activated. It anticipates $A^{(2)}$ and $B^{(3)}$ to occur for the next time-step. \textit{Node} $A^{(2)}$ is not anticipated even if there is a \textit{link} from $X^{(3)}$ to $A^{(3)}$, because the \textit{truth-value} (\textit{i.e.}, the strength of the \textit{link}) is too low. At $t_1$, \textit{concept} $B$ is activated. \textit{Node} $B^{(3)}$ is activated due to the anticipation, while the other $nodes$ in $B$ remain silent. In the meantime, $C^{(3)}$ is anticipated. Similarly, at $t_2$, $C^{(3)}$ is activated, and $Y^{(1)}$ is anticipated.
    (c) The learning mechanism. When there is no \textit{link} between two \textit{concepts}, some \textit{links} are built as \textit{hypotheses} (see Sec.~\ref{sec:hypothesizing}). When one or two \textit{nodes} at both ends of a \textit{link} are activated, the \textit{truth-value} in the \textit{link} is revised according to distinct situations (see Sec.~\ref{sec:revising}). When the number of \textit{links} exceeds a threshold, one or some of them are deleted (see Sec.~\ref{sec:recycling}).
    }
    \label{fig:model}
\end{figure}

\subsection{Representation}\label{sec:repr}

It is explained above why to use such a representation, and the formal description is given in this sub-section.
Generally speaking, the model can be illustrated by two languages, a neural one and a logical one. A more abstract representation, called \textit{Graph representation}, is used to describe the model. The correspondence of the terms with respect to the three representations is shown in Tab.~\ref{tab:terms}.

\subsubsection{Neural Representation}\label{sec:neural-repr}

As we see in Fig.~\ref{fig:model}, there are some \textit{columns} containing several \textit{nodes}. Each \textit{node} represents an artificial (spiking) neuron, which has three possible states, \textit{i.e.}, \textit{resting} (or \textit{non-active}) state, \textit{depolarized} (or \textit{predictive}) state, and \textit{active} state\footnote{An elaborate model of spiking neuron with other states is much more complex, but here is the simplified spiking neuron which contains necessary parts for the sequence learning model.}, 
the meanings of which are similar to those in spiking neurons~\cite{gerstner2014snn}. A neuron in the \textit{predictive} or \textit{resting} state transfers into the \textit{active} state once it is stimulated to some degree, while one in the \textit{resting} state turns into \textit{predictive} state if stimulated. In this paper, the former case is called that a neuron is \textit{pre-activated} for short, and latter case is that a neuron is \textit{activated}. Inspired by HTM~\cite{hawkins2016seq-mem}, the model here owns the structure of \textit{mini-column} and follows the same activation rule as in HTM. 

A \textit{mini-column} is a special structure in neocortex~\cite{buxhoeveden2002minicolumn}, and it looks like a column of neurons. 
Each neuron in a \textit{mini-column} is activated in a certain or several contexts, as shown in Fig.~\ref{fig:model}.
An \textit{event}'s occurring corresponds to the activation of a \textit{mini-column}. When activating a \textit{mini-column}, if all the \textit{neurons} within it are \textit{non-active}, then all the \textit{neurons} should be activated, meaning that all possible contexts are concerned. In contrast, if there are any neurons in \textit{predictive} state, then only these neurons becomes \textit{active} while remaining other neurons to be \textit{non-active}, meaning that certain contexts occur. 

Formally, suppose the $i$th neuron in \textit{mini-column} $c$, in which there are $n_c$ \textit{neurons}, is denoted by $N_c^{(i)}$. The rule of activating \textit{mini-column} $c$ is shown in Eq.~\ref{eq:activation},
\begin{equation}
\label{eq:activation}
    A_c^{(i)} = \left\{
    \begin{aligned}
        1 & ~\text{if} & \forall j\in \{1,...,n_c\}, \hat A_c^{(j)}=0  \\
        1 & ~\text{if} & \hat A_c^{(i)}=1 \\
        0 & ~\text{if} & \hat A_c^{(i)}=0, \text{and}~ \exists j\in \{1,...,n_c\}, \hat A_c^{(j)}=1  \\
    \end{aligned}
    \right.
\end{equation}
where $A_c^{(i)}$ indicates the \textit{active}/\textit{resting} state of neuron $N_c^{(i)}$ ($A_c^{(i)}=1$ if in \textit{active} state, and $A_c^{(i)}=0$ if in \textit{resting} state), while $\hat A_c^{(i)}$ indicates the \textit{depolarized} state of neuron $N_c^{(i)}$ ($\hat A_c^{(i)}=1$ if in \textit{depolarized} state, otherwise, $\hat A_c^{(i)}=0$).

Neurons are connected by synapses. Each synapse is plastic, meaning that its strength is changeable. Some learning rules regarding synapse were proposed, such as \textit{Hebbian} learning, STDP,~\cite{gerstner2014snn}\textit{etc}. If the strength is greater than a threshold (denoted as $\theta$ here), pre-synaptic neuron's activation would lead to post-synaptic neuron's activation, otherwise, the synapse is a \textit{potential} connection waiting for being strengthened. The depolarization procedure is expressed by Eq.~\ref{eq:depolarization},
\begin{equation}
\label{eq:depolarization}
    \hat A_{c_1}^{(i),t} = \left\{
    \begin{aligned}
        1 & ~\text{if}~ A_{c_2}^{(j),t-1}=1 ~\text{and}~ W_{c_1(i)}^{c_2(j)}>\theta \\
        0 & ~\text{otherwise}
    \end{aligned}
    \right.
\end{equation}
where $\hat A_{c_1}^{(i),t}$ indicates the depolarized state of neuron $N_{c_1}^{(i)}$ at time-step $t$,  $A_{c_2}^{(j),t-1}=1$ denotes the active state of neuron $N_{c_2}^{(j)}$ at time-step $t-1$, and $W_{c_1(i)}^{c_2(j)}$ is the strength of the synapse connecting neuron $N_{c_1}^{(i)}$ to neuron $N_{c_2}^{(j)}$.

The rule of modifying synaptic strength is not explicitly presented here. Generally speaking, it is similar to \textit{Hebbian} rule: a synapse is strengthened if its pre-synaptic and post-synaptic neurons are activated simultaneously, and is weakened if only one of the neurons is activated in a short duration. The learning rule in this paper is a variant of \textit{Hebbian} rule (see Sec.~\ref{sec:logic-repr} and Sec.~\ref{sec:seq-learning}).

Different from the HTM theory, in which an event is represented by \textit{sparse distributed representation}\footnote{Briefly speaking, a \textit{sparse distributed representation} in HTM is a binary vector with a little amount of elements to be $1$ and the others to be $0$.}~\cite{hawkins2016bami}
, the model in this paper adopts \textit{concept-centered} representation (see \ref{sec:logic-repr}), so that the model can work in a human-understandable way.

\subsubsection{Logical Representation}\label{sec:logic-repr}

The logical representation in this paper is \textit{concept-centered}, meaning that an \textit{event} is represented by a single \textit{concept} (\textit{i.e.}, a single \textit{mini-column} instead of a set of \textit{mini-columns} as in HTM).

In sequence learning, a representation should be highly contextual. However, despite of the biological-plausibility and robustness, there seems to be no strong reason why \textit{sparse distributed representation} (SDR) is necessary for intelligence. In  the meanwhile, how to deal with uncertainty is a challenge in HTM~\cite{hawkins2016seq-mem}. In principle, a collection of \textit{neurons} in SDR is equivalent to a \textit{concept} in Non-Axiomatic Logic (NAL)~\cite{wang2013nal} in some sense. I believe the most critical intuition in HTM is that a \textit{mini-column} involves a collection of representations under multiple contexts, and each \textit{neuron} in a \textit{mini-column} is highly related to a certain context. It is natural to think if we could use a single \textit{neuron} or \textit{mini-column}, instead of multiple ones, as a representation, simultaneously preserving the intuition in HTM.
From another perspective, each \textit{concept} in Non-Axiomatic Reasoning System (NARS)~\cite{hammer2016opennars}, an AGI system based on NAL, is \textit{weakly contextual}, meaning that what \textit{concept} to be activated is determined by the overall status of the system, while it is not determined directly by what is activated at present. As a result, it is possible that that the current NARS is a good model of consciousness~\cite{wang2020consciousness}, however, it still needs to be improved for \textit{sequence learning}. By exploiting the logic part of NARS, namely NAL, the model proposed can work with uncertainty, and new representations can be derived via well justified logical rules, promising the interpretability of the model.

The schematic diagram of the representation approach is shown in Fig. \ref{fig:model}a. A \textit{column} is interpreted as a \textit{concept}. Within each column, there are several \textit{nodes}. A \textit{node} is interpreted as a \textit{task} that is comprised of a \textit{statement}, a \textit{budget}, and a \textit{truth-value}. \textit{Statement} is the identity of \textit{task}. A \textit{task}'s occurring means that the agent is perceiving or feeling something at a certain time. For example, when seeing a red flower, a \textit{task} which represents the red flower raises up, in other words, it feels the red flower.  \textit{Truth-value}, which represents the extent of the agent's perceiving or feeling, is composed of two parts, \textit{frequency} (denoted as $f$) and \textit{confidence} (denoted as $c$), represented by a two-dimensional tuple $\langle f; c \rangle$. \textit{Frequency} measures ratio of \textit{positive evidence} among all observations, and \textit{confidence} reflects the impact of future evidence\footnote{In NAL, there is no ``absolute truth'', and the truth of a judgement is evaluated by the evidence the system has observed. Suppose there are $w^+$ pieces of positive evidence and $w^-$ negative evidence, then the total amount of evidence is $w=w^++w^-$. \textit{Frequency} is measured by $f=w^+/w$, while \textit{confidence} is measured by $c=w/(w+k)$, where $k$ is a constant.}. 
Both $f$ and $c$ indicates the uncertainty of a \textit{statement}.
A \textit{node}, as a \textit{task}, is also an \textit{event} in NAL since its \textit{truth-value} is time-dependent. \textit{Budget} represents the extent of computation resources allocated to a \textit{task}; it is highly related to an agent's attention. An \textit{event}, in this sense, is not what occurs outside the mind but the subjective experience of the occurrence. Even though a single \textit{concept} corresponds to multiple \textit{events}, under a certain context, usually there should be only one or very few \textit{events} to be activated, so that only part of \textit{meanings} of the concept is utilized.

The temporal relations between two concepts $E_1$ and $E_2$ include predictive implication ``$\langle E_1 \predictiveImplication E_2 \rangle$'', retrospective implication ``$\langle E_2 \retrospectiveImplication E_1\rangle$'', and predictive equivalence ``$\langle E_1 \predictiveEquivalance E_2\rangle$''. 
A sequence of events can be represented as ``$(E_1, E_2, ..., E_n)$''.

As shown in Fig.~\ref{fig:model}a, a chain of nodes represents multiple beliefs simultaneously. For example , ``$\langle A^{(1)} \predictiveEquivalance B^{(1)} \rangle$'', ``$\langle (A^{(1)}, B^{(1)}) \predictiveEquivalance C^{(3)} \rangle$'', and ``$\langle (A^{(1)}, B^{(1)}, C^{(3)}) \predictiveEquivalance D^{(4)} \rangle$'' shares the same chain. 

Given two events $E_1.~\langle f_1; c_1 \rangle$ and $E_2.~\langle f_2; c_2\rangle$, and their corresponding occurrence time $t_1$ and $t_2$ such that $t_2 < t_1$, the temporal induction rules in NAL includes
\begin{align}
\label{eq:induction}
    \{ E_1~\langle f_1; c_1 \rangle, E_2.~\langle f_2; c_2 \rangle \} & \vdash E_2 \predictiveImplication E_1~\langle F_{ind} \rangle \\
    \{ E_1~\langle f_1; c_1 \rangle, E_2.~\langle f_2; c_2 \rangle \} & \vdash E_1 \retrospectiveImplication E_2~\langle F_{ind}' \rangle \\
    \{ E_1~\langle f_1; c_1 \rangle, E_2.~\langle f_2; c_2 \rangle \} & \vdash E_2 \predictiveEquivalance E_1~\langle F_{com} \rangle
\end{align}
where $F_{ind}$ and $F_{com}$ are \textit{induction} and {functions} which map the truth-values of premises to that of conclusion\footnote{In $\langle F_{ind} \rangle$, $w^+=f_1f_2c_1c_2$ and $w=f_2c_1c_2$; In $\langle F_{ind}' \rangle$, $w^+=f_1f_2c_1c_2$ and $w=f_1c_1c_2$,; in $\langle F_{com} \rangle$, $w^+=f_1 f_2 c_1 c_2$, $w=(1-(1-f_1)(1-f_2))c_1c_2$. \textit{Frequency} and \textit{confidence} are then calculated by $f={w^+}/{w}$ and $c={w}/{w+k}$.}. 

For each event, the truth-value is constant (\textit{e.g.}, $\langle 1.0;0.9 \rangle$) in this paper, though they could be revised dynamically in future work. The \textit{truth-value} of an anticipation can be derived by temporal \textit{deduction} rule in NAL, for example,
\begin{equation}
    \{ E_1~\langle f_1; c_1 \rangle, E_1 \predictiveEquivalance E_2.~\langle f_2; c_2 \rangle \} \vdash E_2~\langle F_{ded} \rangle
\end{equation}
where $F_{ded}$ is \textit{deduction function}\footnote{In $\langle F_{ded} \rangle$, $f=f_1 f_2$ and $c = f_1 f_2 c_1 c_2$}. When a concept is activated, that is, a corresponding event ``$E. \langle f;c \rangle$'' occurs, all the contextual-concepts ``$E^{(i)} \langle f^{(i)};c^{(i)} \rangle$'' ($i\in {1,...,n}$, where $n$ is the number of contextual-concepts of concept $E$) are observed as a fact represented by a \textit{truth-value}, so that the \textit{revision} rule is applied to merge the two \textit{truth-values} of fact and anticipation. The \textit{revision} rule in NAL is
\begin{equation}
\label{eq:revision}
    E \langle f_1;c_1 \rangle, E \langle f_2;c_2 \rangle \vdash E \langle F_{rev} \rangle
\end{equation}
where $F_{rev}$ is \textit{revision function}\footnote{In $\langle F_{rev} \rangle$, $w^+=w_1^+ + w_2^+$, $w^-=w_1^- + w_2^-$, and $w=w^+ + w^-$}. It is implied that an anticipated event has higher $confidence$ when it actually occurs.

When a \textit{concept} is activated, which \textit{contextual-concept} to be activated depends on the \textit{expectations} of the truth-values. In NAL, expectation of statement  ``$S \langle f;c \rangle$'' is
\begin{equation}
\label{eq:expectation}
    e(S) = F_{exp}(f,c) = c(f-0.5)+0.5
\end{equation}
We can see that there exists such a threshold $\zeta$, such that the \textit{expectation} of both an occurring but not anticipated \textit{event}, or an anticipated but not occurring \textit{event}, is less than $\zeta$, while the \textit{expectation} of an occurring and anticipated event is greater than $\zeta$. 
A \textit{contextual-concept} is activated when its \textit{expectation} is greater than $\zeta$, or when all the \textit{expectations} of \textit{contextual-concepts} in a \textit{concept} are less than $\zeta$, \textit{i.e.}, 

\begin{equation}
\label{eq:activation2}
    A_c^{(i)} = \left\{
    \begin{aligned}
        1 & ~\text{if} & \forall j\in \{1,...,n_c\}, e(E_c^{(j)})<\zeta  \\
        1 & ~\text{if} & e(E_c^{(j)})>\zeta \\
        0 & ~\text{if} & e(E_c^{(i)})<\zeta, \text{and}~ \exists j\in \{1,...,n_c\}, e(E_c^{(j)})>\zeta  \\
    \end{aligned}
    \right.
\end{equation}

The procedure of temporal induction for statement ``$E_{c_1}^{(i)} \predictiveEquivalance E_{c_1}^{(j)}$'' happens only when $A_{c_1}^{(i)}=1$ or $A_{c_2}^{(j)}=1$. 

Although predictive implication (``$\predictiveImplication$'') and retrospective implication (``$\retrospectiveImplication$'') are also important, as a start point, the model in this paper exploits merely predictive equivalence (``$\predictiveEquivalance$'') for learning and inference.

\subsubsection{Graph Representation}

We have seen in Sec. \ref{sec:neural-repr} and Sec.~\ref{sec:logic-repr} that the model can not only be explained as a neuronal network, but also be interpreted by a logic. However, to better illustrate it, I have to use a more abstract representation (which is called \textit{Graph Representation}\footnote{This might not be a suitable term, but I was not able to find a better one.}) to avoid conceptual ambiguity in the description, mainly by using the formal language of \textit{Graph Theory}. As shown in Fig.~\ref{fig:model}, the basic elements are \textit{node} and \textit{link} (\textit{a.k.a.} vertex and directed edge in \textit{Graph Theory}). A \textit{column} (as hyper-vertex) is a collection of \textit{nodes}. Each \textit{link}'s weight is adjustable. Each \textit{node} has three states, \textit{activation}, \textit{non-activation}, and \textit{pre-activation}, each of which is represented by a pair of real numbers (\textit{i.e.}, \textit{truth-value} in Sec.~\ref{sec:logic-repr}) ranging from 0 to 1. The states of real number can be binarized by a threshold (see Eq.~\ref{eq:activation2}).

The correspondence among the terms in the three representations is shown in Tab.~\ref{tab:terms}.

\begin{table}[h]
    \centering
    \begin{tabular}{c|c|c}
    \toprule
        Graph Repr. & Neural Repr. & Logical Repr. \\
    \midrule
        node & neuron & contextual-concept \\
        column & mini-column & concept \\
        link & synapse & temporal statement (\textit{e.g.}, ``$\langle A\predictiveEquivalance B\rangle$'') \\
        link-weight & synaptic strength & truth-value (\textit{abbr.}, t.-v.) \\
        weight adjustment & synaptic plasticity & temporal-induction and revision \\
        activation & active state & event with t.-v. ``$(1.0;0.9)$''\tnote{*1} \\
        pre-activation & depolarized state & anticipation with t.-v. ``$(1.0;0.9)$''\tnote{*1} \\
        non-activation & resting state & event with t.-v. ``$(1.0;0.1)$''\tnote{*2} \\
    \botrule
    \end{tabular}
    \begin{tablenotes}
        \footnotesize
        \item[*1] Here in the truth-value, \textit{frequency} and \textit{confidence} are both very high, though the concrete values does not have to be the same as $(1.0;0.9)$.
        \item[*2] Here in the truth-value, \textit{confidence} is very high but \textit{frequency} does not matter, and the concrete values does not have to be the same as $(1.0;0.1)$.
    \end{tablenotes}    
    \caption{The correspondence of terms among the three representations: \textit{Neural Representation}, \textit{Logical Representation}, and \textit{Graph Representation}.}
    \label{tab:terms}
\end{table}

\subsection{Sequence Learning}
\label{sec:seq-learning}

The challenge is how to construct the \textit{links} given a series of \textit{events}. First, there is no way to fully-connect among all nodes (meaning that a node has links connected to all other nodes). Facing an endless list of events, the number columns cannot be pre-determined (thus, new columns should be able to be built up dynamically), and the number of links would explode as the number of columns increases with fully-connection. Due to the Assumption of Insufficient Knowledge and Resources (AIKR) \cite{wang2019defining}, the number of links connected to or from a node should not exceed constant (though it could be either large or small), consequently, there has to be a certain mechanism through which new \textit{links} are created with old \textit{links} to be recycled. In Sec.~\ref{sec:logic-repr}, the logic rules of temporal \textit{induction} has been introduced, however, when to do induction and to revise the link remains to be answered in the following.

\subsubsection{Hypothesizing}\label{sec:hypothesizing}

Initially, there are no \textit{nodes} and no \textit{links} in the network. Whenever an \textit{event} occurs, the corresponding \textit{column} is constructed if there does not exist one. Each \textit{node} in a \textit{column} has no links at the beginning of its creation. When two \textit{columns} are activated in succession, two sets of nodes are activated correspondingly. Suppose a set of nodes $\mathcal{N}_1$ in column ${C}_1$ and a set of nodes $\mathcal{N}_2$ in column ${C}_2$ are activated, then one node $E_{c_1}^{(i)}$ is picked out from $\mathcal{N}_1$, and another one $E_{c_2}^{(j)}$ from $\mathcal{N}_2$, a new link is created connecting from $E_{c_1}^{(i)}$ to $E_{c_2}^{(j)}$ if there does not exist one. Since the initial weight of the link, represented by truth-value, is very weak, \textit{i.e.}, the \textit{confidence} is low (\textit{e.g.}, $c=0.1$). The link, represented by ``$E_{c_1}^{(i)} \predictiveEquivalance  E_{c_2}^{(j)} \langle 1.0; 0.1 \rangle$'', in this sense is what we usually mean by \textit{hypothesis}.

When picking out a node for hypothesizing from a set, which one to pick? Let us consider the \textit{meaning} of a node. A node is activated given a context of events, thus, intuitively, a node means a concept in a certain context. Ideally, there should be at most one \textit{pre-link} pointing into it and at most one \textit{post-link} pointing out from it, and there should at least one link possessed by it. In this case, the node would be activated if and only if one certain context of events occur. For example, a node $B^{(1)}$ is activated only when the sequence ``$(A, B, C, D)$'' occurs, and $B^{(1)}$ exactly identifies the event $B$ in that context, rather than the $B$ in ``$(X, B, C, Y)$''. However, due to AIKR, the number of nodes should be a constant, so that a node has to serve for multiple different contexts. We can say the meaning of a node is \textit{clear} or \textit{unambiguous}, \textit{either} if it has a post-link with strength much greater than other post-links, and a pre-link with strength much greater than other pre-links, \textit{or} if one of its pre-link and post-link is much greater than its other links. To pick out one for hypothesizing, the overall principle is to avoid as far as possible to do harm to the \textit{clear} meaning of a node. The concrete strategy in this paper is to pick out a node with the lowest \textit{utility}. Here, the utility of node $E_c^{(i)}$ is defined as

\begin{equation}
    u(E_c^{(i)}) = 1-(1-u_1)(1-u_2)
\end{equation}
where
\begin{equation}
\begin{aligned}
    u_1 & = \left\{ \begin{aligned}
        &0 &,&~\text{if}~\mathcal{L}_{pre}(E_c^{(i)}) = \varnothing \\
        &\max_{\forall L \in \mathcal{L}_{pre}(E_c^{(i)})} e(L) &,& ~\text{otherwise}
    \end{aligned}  \right. , ~\text{and} \\
    u_2 & = \left\{ \begin{aligned}
        &0 &,&~\text{if}~\mathcal{L}_{post}(E_c^{(i)}) = \varnothing \\
        &\max_{\forall L \in \mathcal{L}_{post}(E_c^{(i)})} e(L) &,& ~\text{otherwise}
    \end{aligned}  \right.
\end{aligned}
\end{equation}
where $\mathcal{L}_{pre}(E_c^{(i)})$ and $\mathcal{L}_{post}(E_c^{(i)})$ are the sets of node $E_c^{(i)}$'s pre-links and post-links correspondingly, and $e(L)$ is the \textit{expectation} of the truth-value of link $L$ (see Eq.~\ref{eq:expectation}). Thus, if a node has a much clear meaning, it tends not to be picked out. Though the side effect is that a node, which has ambiguous meaning but has a link with strong strength, is also inclined to be selected, it seems not an issue in practice.

New \textit{hypotheses} are constantly come up with, though they do not lead to strong conclusion until enough evidences are collected. A too weak hypothesis like ``$E_{c_1}^{(i)} \predictiveEquivalance  E_{c_2}^{(j)} \langle 1.0; 0.1 \rangle$'' leads to non-activation of its consequent $E_{c_2}^{(j)}$, according to Eq.~\ref{eq:activation2}, in this sense, a hypothesis is a \textit{potential} link between two nodes. This potential link is similar to a synapse with a low strength. Only when the strength is greater than a threshold, the synapse can be viewed as truly connected and can transit signals between two neurons. Nevertheless, a link can be strengthened whether it is strong or weak, so that a \textit{hypothesis} has chance to become stronger and cause the activation of its consequent given the activation of its antecedent.

\subsubsection{Revising}\label{sec:revising}

Whenever a column is activated, one link is picked out for revising. The general principle to enhance the link is that is the most probable to become conclusive. Specifically, the selected link $L$ is 
\begin{equation}
    L=\underset{{\forall L\in\mathcal{L}_{pre}(E_{c}^{(i)}),\forall i \in \{1,...,n_c\}}}{\mathrm{argmax}}e(L)
\end{equation}
meaning that it picks out a link with the maximal \textit{expectation} from all the pre-links of all the nodes within a column. As a result, a link represented by statement ``$E_{c_1}^{(i)} \predictiveEquivalance E_{c_2}^{(j)}$'' is selected for revising.

Given two nodes $E_{c_1}^{(i)}$ and $E_{c_2}^{(j)}$ which are concerned on, the temporal induction rule is applied to revise the truth-value of statements including ``$E_{c_1}^{(i)} \predictiveEquivalance E_{c_2}^{(j)}$'', according to Eq.~\ref{eq:induction} and Eq.~\ref{eq:revision}. The difference in the learning procedure is that some negative evidences are obtained if an anticipated event does not occur. Specifically, node $E_{c_2}^{(j)}$ (as consequent) is in the state of \textit{pre-activation} at time-step $t$ (\textit{i.e.}, $\hat{A}_{c_2}^{(i), t}=1$) if  node  $E_{c_1}^{(i)}$ (as antecedent) is in the state of \textit{activation} at time-step $t-1$ (\textit{i.e.}, $A_{c_1}^{(i),t-1}=1$), and the expectation of statement ``$E_{c_1}^{(i)} \predictiveEquivalance E_{c_2}^{(j)}$'' is greater than a threshold $\theta$, \textit{i.e.},
\begin{equation}
    \hat{A}_{c_2}^{(i), t} = \left\{ \begin{aligned}
        &1&,&~\text{if}~ A_{c_1}^{(i),t-1}=1, ~\text{and}~ e(E_{c_1}^{(i)} \predictiveEquivalance E_{c_2}^{(j)}) > \theta &\\
        &0&,&~\text{otherwise}&
    \end{aligned} \right.
\end{equation}
An event usually may have multiple causes and effects, however, in the sequence learning model, It is possible to build \textit{causal chains} in which each event has at most only one cause and at most one effect in a certain context. Therefore,  in the case of $\exists j\in \{1,...,n_c\}, e(E_c^{(j)})>\zeta$ (\textit{i.e.}, the current event has a certain context, such that one or some of nodes within a column are activated but not all), if node $E_{c_2}^{(j)}$ is anticipated ($\hat{A}_{c_2}^{(i), t}=1$) but not activated (${A}_{c_2}^{(i), t}=0$), then some negative evidences of  $E_{c_2}^{(j)}$ are collected. Similarly, when node $E_{c_2}^{(j)}$ is activated, all of its possible causes are paid attention to. If an antecedent $E_{c_1}^{(i)}$ is not activated before $E_{c_2}^{(j)}$, given statement ``$E_{c_1}^{(i)} \predictiveEquivalance E_{c_2}^{(j)}$'', then some negative evidences of $E_{c_2}^{(j)}$ are also collected. Otherwise, the two nodes are activated in succession, temporal induction can be applied directly. \footnote{In practice, A simplified but equivalent implementation is adopted. If nodes $E_{c_1}^{(i)}$ and $E_{c_2}^{(j)}$ are activated in succession, then an amount of positive evidences, $w^+=p^+$, are collected for statement ``$E_{c_1}^{(i)} \predictiveEquivalance E_{c_2}^{(j)}$''. However, if only one of the nodes is activated,  then some negative evidences , $w^-=p^-$ are collected. Here constants $p^+$ and $p^-$ are hyper-parameters of the model, typically $p^+=p^-=1$.}

There is also a punishment in the case $\forall j\in \{1,...,n_c\}, e(E_c^{(j)})<\zeta$ (\textit{i.e.}, all the nodes within a column are activate without being anticipated first). If the whole column is activated, a bunch of anticipations would occur. For those anticipations which are not verified later, a slight amount of evidences are collected by $w^-=|\{e(L)>\theta,\forall L\in\mathcal{L}_{post}(E_{c_1}^{(i)})\}|+b$, where the first term denotes the number of $E_{c_1}^{(i)}$'s post-links each of whose \textit{expectation} is greater than threshold $\theta$, and the second term  $b$ is a constant (\textit{e.g.}, $b=40$); here, threshold $\theta$ and constant $b$ are hyper-parameters of the model. When $b=\infty$, it means no penalty for this case.

\subsubsection{Recycling}\label{sec:recycling}

Again due to AIKR, the number of links regarding a node should not exceed a certain threshold, otherwise, some of the links should be dropped. This is related to the forgetting process of memory.

\textit{Pre-links} $\mathcal{L}_{pre}(E_{c_1}^{(i)})$ and \textit{post-links} $\mathcal{L}_{post}(E_{c_1}^{(i)})$  of node $E_{c_1}^{(i)}$ are stored in a priority queue, sorted by \textit{utility}. In the current design, \textit{utility} of a \textit{link} is determined by the \textit{expectation} of its \textit{truth-value}. When the number of \textit{link} $n_L$ in the priority queue is greater than a certain threshold $\xi$ (\textit{e.g.}, $\xi=100$), the exceeding part is recycled, \textit{i.e.}, $(n_L-\xi)$ links with the lowest priorities are deleted.

\textit{Utility} probably not only depends on the \textit{expectation} of a \textit{link}'s \textit{truth-value}, but also some other factors. For example, a link may be reserved in a short period after it is newly created, though its \textit{expectation} is much less than some long-standing links (see \textit{Future Work} for more discussion in Sec.~\ref{sec:future}).

\section{Results}\label{sec_results}

Since the tasks, sequence \textit{prediction} and sequence \textit{generalization}, are equivalent to each other, while the sequence \textit{recognition} task can be reduced to \textit{prediction}\cite{sun2001sequence}, the test-cases in this paper only involves sequence \textit{prediction}. In the meanwhile, sequence \textit{decision making} task~\cite{sun2001sequence} can be considered as associated with much more complex procedures of intelligence, thus, \textit{decision making} is not considered in this paper, though the model  proposed in this paper is the foundation of further work.

The model is tested on some synthetic datasets. Different from typical approach for evaluating machine learning methods, there is no explicit division between training set and test set here, since it is considered that the learning process is \textit{online} and \textit{life-long}: each sample observed by an agent is not only a training sample but also a test one, and current experience is not necessary to be similar to the past, that is to say, no stable distribution of data is assumed.

The data for tests are manufactured in the following way. Suppose there are $n_r$ types of events; each event is labeled by a \textit{term}, such as characters like ``$A$'', ``$B$'', ``$X$'', and strings like ``$\textit{e0347}$'', ``$\textit{e1001}$'', which identifies the type of an \textit{event}. Whatever a \textit{term} looks like for human developers, it is just the name of a \textit{concept} inside the system, a \textit{concept} whose meaning merely depends on its acquired (rather than predetermined) relations with other \textit{concepts}, as suggested in Non-Axiomatic Logic (NAL)~\cite{wang2013nal}. A dataset in this paper is a list of \textit{events}, which contains sequences like ``$(...,A, B, C, D,...)$'', ``$(...,X, B, C, Y,...)$'', and so on. Some \textit{events} are determined by their predecessors, for example, in a given situation, \textit{event} ``$B$'' is always followed by ``$C$'' but comes after either ``$A$'' or ``$X$''; event ``$D$'' follows ``$(A,B,C)$'', but given merely ``$(B,C)$'', either ``$D$'' or ``$Y$'' is expected to occur. Besides, other events are randomly generated, leading to the whole list of events unpredictable to some extent.

With this form of input data, three aspects are considered for evaluating the model, \textit{capacity} (see Sec.~\ref{sec:capacity-tests}), \textit{catastrophic forgetting} (see Sec.~\ref{sec:forgetting-test}), and \textit{capability} (see Sec.~\ref{sec:capability-test}), though the capability aspect is analyzed only in theory.

\subsection{Capacity Tests}\label{sec:capacity-tests}

Evaluating the capacity of the model is related to two factors, the number of sequences and the length of a sequence that is expected to be recognized. 
In a test, datasets are generated, within the prototype of ``$(\$, ..., \$, E_1, ..., E_m, \$, ..., \$)$'', where $E_1, ..., E_m$ are deterministic events which keep the same in every sample of the prototype, while ``$\$$'' is the unpredictable variable, which varies from sample to sample; The dataset's parameter $m$ denotes the number of deterministic events for each sequence. There are $p$ pieces of prototypes of sequences to be generated. When an event occurs, the model anticipates some events to occur for the next step. If an event is anticipated and occurs immediately, then we can say the event is correctly anticipated. The proportion of the number of truly anticipated events within a certain past period (\textit{e.g.}, the past 100 time-steps) is the anticipation accuracy of the current time-step. 

\begin{figure}[h]
\centering
\begin{subfigure}{0.32\textwidth}
    \includegraphics[width=\textwidth]{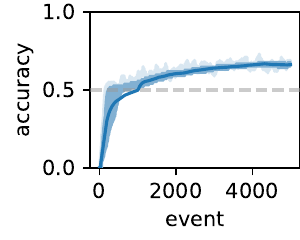}
    \caption{}
    \label{fig:cap-simple-case}
\end{subfigure}
\hfill
\begin{subfigure}{0.32\textwidth}
    \includegraphics[width=\textwidth]{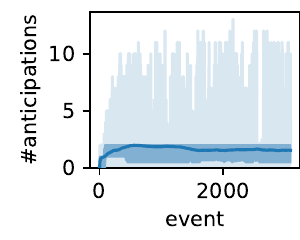}
    \caption{}
    \label{fig:cap-simple-case-anticipations}
\end{subfigure}
\hfill
\begin{subfigure}{0.32\textwidth}
    \includegraphics[width=\textwidth]{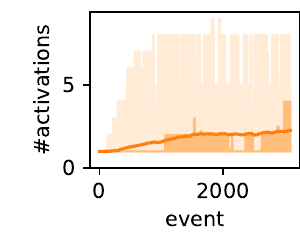}
    \caption{}
    \label{fig:cap-simple-case-activations}
\end{subfigure}

\caption{Capacity-Test results for the simple case, where the prototypes of sequences are  ``$(\$, \$, A, B, C, D, E, \$)$'' and ``$(\$, \$, X, B, C, D, Y, \$)$'', where ``$\$$'' denotes a random event. (a) The accuracy of anticipation. (b) The number of anticipations. (c) The number of active nodes.}
\label{fig:capacity-tests-simple-case}
\end{figure}

Firstly, a simple case is tested. Suppose each event is named by a single character (from $A$ to $Z$), so that there are $26$ possible types of events for the model. The dataset contains two prototypes of sequences ``$(\$, \$, A, B, C, D, E, \$)$'' and ``$(\$, \$, X, B, C, D, Y, \$)$'', where ``$\$$'' denotes a random event. In this case, $m=5$ and $p=2$), and only $50\%$ of the events are deterministic and can be predicted very well.  The test results are shown in Fig.~\ref{fig:capacity-tests-simple-case}. Figure \ref{fig:cap-simple-case} shows the accuracy of anticipation as time goes by. At each time-step, there could be multiple anticipations, and Fig.~\ref{fig:cap-simple-case-anticipations} shows the number of events anticipated by the model -- ideally, there should be only one anticipated event if the system is pretty sure what context it observes; multiple anticipated events implies that the system retains the possibility of several contexts. We can see that around 2 events are anticipated on average for each time-step. Figure~\ref{fig:cap-simple-case-activations} shows the number of activated nodes in the model -- generally speaking, a node's activation means a certain context is recognized by the model. The fewer nodes are active, the clearer the context is. It shows in Fig.~\ref{fig:cap-simple-case-activations}, there are around $2$ nodes activated on average for each time-step.

\begin{figure}
\centering
\begin{subfigure}{0.32\textwidth}
    \includegraphics[width=\textwidth]{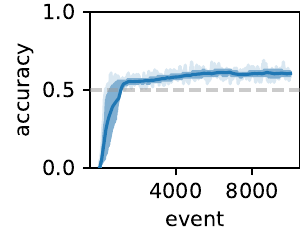}
    \caption{}
    \label{fig:cap-5-5}
\end{subfigure}
\hfill
\begin{subfigure}{0.32\textwidth}
    \includegraphics[width=\textwidth]{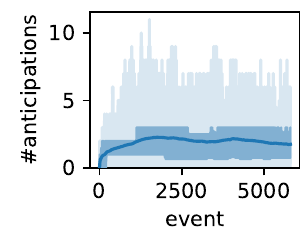}
    \caption{}
    \label{fig:cap-5-5-anticipations}
\end{subfigure}
\hfill
\begin{subfigure}{0.32\textwidth}
    \includegraphics[width=\textwidth]{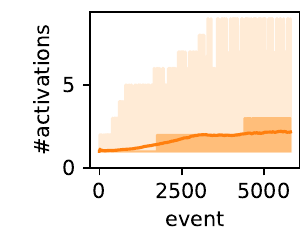}
    \caption{}
    \label{fig:cap-5-5-activations}
\end{subfigure}
\hfill
\begin{subfigure}{0.32\textwidth}
    \includegraphics[width=\textwidth]{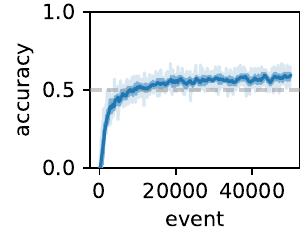}
    \caption{}
    \label{fig:cap-14-20}
\end{subfigure}
\hfill
\begin{subfigure}{0.32\textwidth}
    \includegraphics[width=\textwidth]{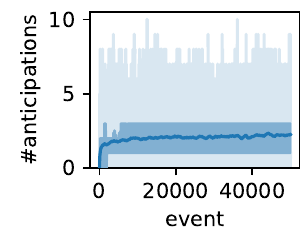}
    \caption{}
    \label{fig:cap-14-20-anticipations}
\end{subfigure}
\hfill
\begin{subfigure}{0.32\textwidth}
    \includegraphics[width=\textwidth]{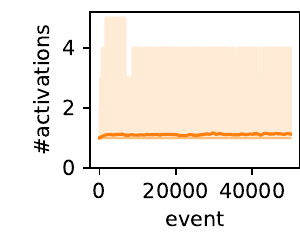}
    \caption{}
    \label{fig:cap-14-20-activations}
\end{subfigure}

\hfill
\begin{subfigure}{0.32\textwidth}
    \includegraphics[width=\textwidth]{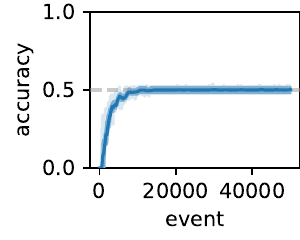}
    \caption{}
    \label{fig:cap-14-1000}
\end{subfigure}
\hfill
\begin{subfigure}{0.32\textwidth}
    \includegraphics[width=\textwidth]{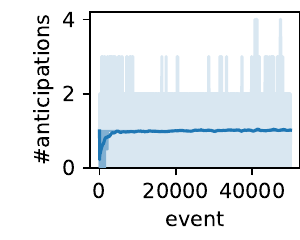}
    \caption{}
    \label{fig:cap-14-1000-anticipations}
\end{subfigure}
\hfill
\begin{subfigure}{0.32\textwidth}
    \includegraphics[width=\textwidth]{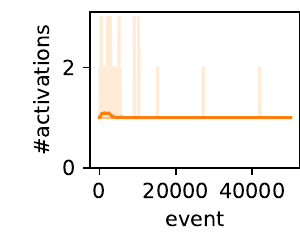}
    \caption{}
    \label{fig:cap-14-1000-activations}
\end{subfigure}

\caption{Capacity-Test results with different options of length $m$ and the number of  prototypes $p$, and event the different numbers of types of evens. (a), (b), and (c) are test results with $m=5$, $p=5$, and $26$ types of events. (d), (e), and (f) are test results with $m=14$, $p=20$, and $26$ types of events. (g), (h), and (i) are test results with $m=14$, $p=20$, and $1000$ types of events.  }
\label{fig:capacity-tests}
\end{figure}

Secondly, the model is tested with different options of length $m$ and the number of  prototypes $p$, and even the different numbers of types of events. The proportions of unpredictable events in the datasets are all $50\%$.
As shown in Fig.~\ref{fig:capacity-tests},  the model has proper anticipations on future events. With $m=5$ and $p=5$ (see Fig.~\ref{fig:cap-5-5}), as well as $m=14$ and $p=20$ (see Fig.~\ref{fig:cap-14-20}), the anticipation accuracy in either cases is greater than $50\%$, exceeding the theoretically highest accuracy (the same as that shown in Fig.~\ref{fig:cap-simple-case}). This is because The model learns some patterns from the random events.
The number of anticipated nodes and that of active nodes are both no more than $2$ in each of the two cases.

A probably simpler setting for the model is that the number of types of events is much greater than $26$. The test results are shown in Fig.~\ref{fig:cap-14-1000}-\ref{fig:cap-14-1000-activations}, where the number of types $n_r$ is $1000$. We can see that the the accuracy is closely around $50\%$, and the number of anticipated nodes and that of active nodes are both around $1$. The model performs better in this setting than the previous ones, because in the previous tests, one type of event most probably engages in multiple prototypes of sequences, so that the model may be confused; while in this test, the types of events are much greater, so that one type of event get a higher chance to be  involved in a single context, consequently, it is much easier to memorize and distinguish different patterns for the model.

\subsection{Catastrophic Forgetting Tests}\label{sec:forgetting-test}

The issue of \textit{catastrophic forgetting} (also known as \textit{catastrophic interference}) was proposed by  McCloskey and Cohen in 1989 \cite{mccloskey1989cf}, pointing out that distributed representation of connectionist  networks (\textit{a.k.a.} \textit{Deep Neural Networks} nowadays) have a non-desirable property that modifications on new data interfere the memory for old data, leading to forgetting large amount of the previous experience. Some modern research (\textit{e.g.}, \cite{zeng2019continual}) over the years still tried to solve this issue. 

Since the model in this paper adopts \textit{concept-centered} representation (see Sec.~\ref{sec:logic-repr}), this annoying property seems probably not to occur in theory. However, due to relatively insufficient resources of memory and computation~\cite{wang2019defining} assumed in this paper, the model has to remember something new and forget something old, thus, acquiring new knowledge is possible to interfere old one,
and the extent of interference should be evaluated (at least qualitatively if not quantitatively), to see whether it is catastrophic.

\begin{figure}[h]
    \centering
    \begin{subfigure}{0.64\textwidth}
        \includegraphics[width=1\linewidth]{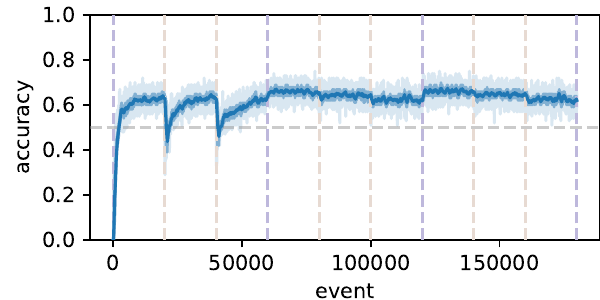}
    \caption{}
    \end{subfigure}
    \  \\
    
        \begin{subfigure}{0.32\textwidth}
            \includegraphics[width=1\linewidth]{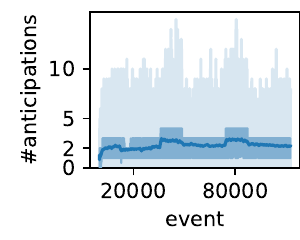}
        \caption{}
        \end{subfigure}
        \begin{subfigure}{0.32\textwidth}
            \includegraphics[width=1\linewidth]{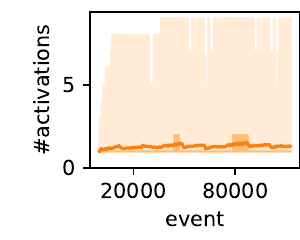}
        \caption{}
    \end{subfigure}
    
    \caption{Results of catastrophic forgetting test when the number of concepts $n_r=26$ (see Sec.~\ref{sec:forgetting-test} for more details).}
    \label{fig:cf-ne=26}
\end{figure}

\begin{figure}
    \centering
    \begin{subfigure}{0.64\textwidth}
        \includegraphics[width=1\linewidth]{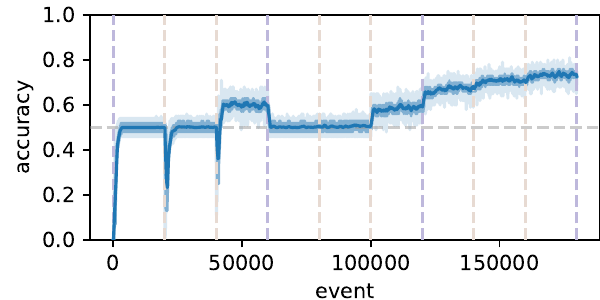}
    \caption{}
    \end{subfigure}
    \  \\
    
        \begin{subfigure}{0.32\textwidth}
            \includegraphics[width=1\linewidth]{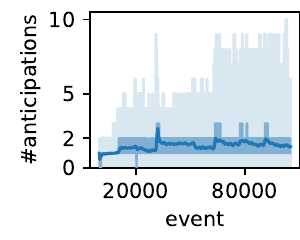}
        \caption{}
        \end{subfigure}
        \begin{subfigure}{0.32\textwidth}
            \includegraphics[width=1\linewidth]{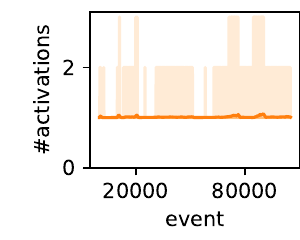}
        \caption{}
    \end{subfigure}
    
    \caption{Results of catastrophic forgetting test when the number of concepts $n_r=1000$ (see Sec.~\ref{sec:forgetting-test} for more details).}
    \label{fig:cf-ne=1000}
\end{figure}

The test results of catastrophic-forgetting is shown in Fig.~\ref{fig:cf-ne=26} and Fig.~\ref{fig:cf-ne=1000}. 
With $26$ types of events (in Fig.~\ref{fig:cf-ne=26}) or $1000$ types (in Fig.~\ref{fig:cf-ne=1000}), in an episode, $20$ prototypes of sequences with length $14$ for each are generated. The patterns vary across different episodes. After seeing three episodes one by one, the model encounters the previous episodes repeatedly. If there existed catastrophic forgetting in the model, then we would have seen the anticipation accuracy fell down significantly when seeing an episode with the same patterns once again. However, that does not happen in Fig.~\ref{fig:cf-ne=26} and Fig.~\ref{fig:cf-ne=1000}.  Therefore,  qualitatively speaking, the model does not suffer from catastrophic forgetting.

\subsection{Capability Analysis}\label{sec:capability-test}

The bound of the model's capability needs to be clarified. 
First, the model is not an AGI system, although it can be considered as a first step on modeling the complex unity of intelligence. 
Second, the model focuses on an aspect of intelligence, \textit{i.e.}, \textit{sequence learning}. 
Third, in the current design, the model can only deal with the situation where merely one single event appears at a certain time-step; the case where multiple events appear simultaneously is not the target in this paper. Forth, the time interval of any pair of events is a constant; this assumption on the interval enables the model to deal with some situations where the order of events matters but time interval does not; an example of this kind of situations is natural language processing. Part of future work is to expand the capability of the model (see \textit{Future Work} in Sec.~\ref{sec_discussion}).

Inside the bound, the model is capable of learning patterns in an endless list of events. In the meanwhile, the model is enabled by the power of Non-Axiomatic Logic to handle uncertainty. This property (\textit{i.e.}, being able to handle uncertainty) is directly derived in theory, thus, no test is needed to prove that in practice, and exploiting the property is more related to applications of the model.

\section{Discussion}\label{sec_discussion}

\bmhead{A Principle}

From the learning mechanism proposed in this paper, the following principle can be summarized:
\begin{quote}
\textit{Computational resources tend to converge toward knowledge with lower levels of uncertainty.}
\end{quote}
Specifically, in the model, a \textit{link} with higher \textit{truth-value} gets a greater chance to be enhanced. This principle is not a novel idea. It could be a guidance for designing AI systems, and it is also observed in biological systems: 
In neuroscience and neuronal dynamics, the well-known \textit{winner-take-all} rule \cite{riesenhuber1999obj-recog, gerstner2014snn, lee1999wta} shares the same intuition. 
In psychology, Piaget's theory suggests that new information input to a subject is incorporated into already existing knowledge\cite{muller2015piaget}. In other words, the existing knowledge will be allocated computational resources to give a meaning to the content.
Earlier, in ancient China, as it is said in \textit{Tao Te Ching}\footnote{A reference translation -- Lao-Tzu, Addiss, S., Lombardo, S., Watson, B.: Tao Te Ching, copyright 1993 edn. Hackett Publishers, Indianapolis (1993).}, ``The \textit{Way of Nature} reduces excess and replenishes deficiency. By contrast, the \textit{Way of Humans} is to reduce the deficient and supply the excessive.'' \footnote{
In Chinese --
\begin{CJK*}{UTF8}{gbsn}
《道德经》云：“天之道，损有余而补不足；人之道则不然，损不足以奉有余。”
\end{CJK*}
} The learning process of the model proposed exactly follows the \textit{Way of Humans}. 

\bmhead{Brain-inspired Structure}

Even if the model is inspired by human brain, in AI systems, it still needs the answer why it should be of the structure like that, rather than a trivial answer as ``because brain looks like that''. The reason why the \textit{mini-column} structure work is that it represents a \textit{concept} with distinct meanings under different contexts. A \textit{neuron} in a \textit{mini-column} is activated because part of the meanings of the corresponding \textit{concept} is recalled. \textit{Neurons} are connected as a \textit{chain}, representing \textit{concepts} organized as a sequence. Due to the Assumption of Insufficient Knowledge and Resources (AIKR), the total number of links should not exceed a constant, so a balance between memorizing and forgetting does matter. Based on this view, the learning mechanism in Sec.~\ref{sec:seq-learning} is designed.

\bmhead{Hierarchy}

We can see that the model is capable of distinguishing the patterns with the same head and tail but different middle parts, \textit{e.g.}, ``$(A, B, C, D)$'' and ``$(X, B, C, Y)$'', however, I do not think it learns a hierarchical structure from a sequence. A pattern is implicitly stored in the memory, in the form of \textit{chain}, rather than \textit{tree} in computer science. ``Chunk'' is the basis of building a hierarchy~\cite{conway2012seq}, and a \textit{chain} could be a hint or heuristic to form a chunk. As suggested in previous work (\textit{e.g.}, \cite{lashley1951hierachy} and \cite{lashley1951hierachy}), forming hierarchical representations of sequences benefits for accessing and self-repairing sequences learned. The model proposed suffers from the problem of memorizing a long sequential pattern: a long \textit{chain} is broken into smaller ones in the learning procedure, as a result, some events in a long sequence cannot be anticipated correctly as expected. This problem, I believe, can be solved by imposing hierarchical structure to the model -- \textit{Nodes} in a \textit{chain} are combined as a \textit{chunk} (i.e., a \textit{compound} in NAL~\cite{wang2013nal}), which simultaneously serves as a \textit{column} and is processed in the same way as what is done in this paper.

\bmhead{Causal Inference}

Due to that temporal implication or equivalence in NAL has the meaning of prediction, a \textit{link} can be viewed as a basic form of causation~\cite{wang2015temporal}, and a \textit{chain} in this paper corresponds to what is called \textit{causal chain}. The model, with no prior knowledge in the beginning, tries to summarize causal relations among events, and those links with high certainty are reserved; this procedure is also called \textit{causal discovery}. Although, by backtracking a \textit{link} or \textit{chain}, causal explanations can be obtained, there are causal explanations that are much more complex in human's daily experience. This issue is too big to be discussed detailly in this paper, nevertheless, I believe this is a novel start point to address the issue of causal inference.

\bmhead{Neuronal Basis of Logic}

How does logic emerge from human brain? There are two possibilities, one is that neural networks might learn something which is called logic, the other is that neuron's activity could be interpreted as logic. Evidently, the work in this paper supports the latter one, though it does not negate the previous one. We can see in Sec.~\ref{sec:repr} and Tab.~\ref{tab:terms} that the model can be illustrated in both two ways, a neural one and a logical one. The correspondence between membrane voltage of a neuron and truth-value of a statement is not discussed deliberately, though I guess there would emerge some valuable work on this issue.

\bmhead{Dropping out the Black-box of Sequence Learning}

This paper proves the potential of learning patterns of sequences based on a logic, the problem which was addressed well by purely statistical models (\textit{e.g.}, Hidden Markov Models~\cite{rabiner1986hmm}), neurodynamics models (e.g., HTM~\cite{hawkins2016seq-mem}, spiking neural networks~\cite{liang2020sl-snn}), and neural networks (\textit{e.g.}, RNN~\cite{medsker2001rnn}). The model is fully interpretable by a logic, Non-Axiomatic Logic~\cite{wang2013nal}, as a result, human developers are capable of explaining the system's behaviors by recording, in a human-understandable way, and checking its internal activities. It provides an alternative besides well-performed but inexplicable models, especially neural networks that are widely criticized as black boxes.

\bmhead{Comparison}

The previous pieces of work are valuable, but there are still some differences from this model.
The practical performances of various models are not compared in this paper, majorly due to two reasons. First, of course, the model proposed here is a preliminary one and is not powerful enough. As suggested above, at least a hierarchy should have been learned by the model to deal with some complex situations. Thus, it does not make much sense to apply it to some complex tasks, such as natural language processing. Second, there are some different theoretical assumptions in this paper. It assumes that the types of \textit{events} are unknown to a system before it is initialized, as a result, corresponding representations should be generated in the run-time. In typical Hidden Markov Models~\cite{rabiner1986hmm}, the types of \textit{events} should be pre-specified and cannot be changed when a system starts running. In neural networks, it usually assumes that data are known satisfactorily to a system, so that the system can see the the whole data set repeatedly. In this work, the assumption is that the data set is endless, thus, the model has to do learn in real time. Besides, the interpretability of the model is an attractive property. The performance of the model proposed is similar to HTM~\cite{hawkins2016seq-mem}, though they adopt quite different theoretical foundations. There are some advantages of distributed representations in HTM, for example, robustness to noise and damage. In contrast, by adopting the \textit{concept-centered} representation, uncertainty can be represented naturally. 

\bmhead{Future Work}\label{sec:future}

There is a great deal of work to be done in the future, related to either improving the sequence learning model \textit{per se} or modeling other intelligence phenomena upon the model proposed.

On one hand, apart from introducing the hierarchical structure mentioned above, the major improvement might involve the competition among links. In the current design, links compete with each other based on its \textit{utility}, however, \textit{utility} of a link should have depend on several factors besides its \textit{expectation}, such as time that a link is established (more specifically, a link built in the recent time tends not to be forgot, even if its \textit{expectation} is low), the goal of a system (therefore, there should be a top-down interaction with the model), and so on. In addition, predictive implication and retrospective implication have not been considered much in the current design, and the learning mechanism could be modified to improve the efficiency of learning. Certainly, the current design of the model is not perfect enough. For example, in Fig.~\ref{fig:cap-5-5-anticipations}, I hope the number of anticipations should be close to $1$, meaning that the system clearly know the context it locates in. The current result shows that there are around $2$ possible contexts, and the system cannot determine which is correct exactly, though it knows to some extent.

On the other hand, sequence learning could be the basis of sensorimotor learning, where the input is high-dimensional data (typically, 2-D image), and an agent perceives its environment by glimpsing different places (\textit{a.k.a.} eye-movement). How to built a interpretable sensorimotor learning model seems a big challenge. Also, multiple events might occur simultaneously, and how to deal with concurrent events deserves further research.










\section{Conclusion}\label{sec13}

In this paper, a model of sequence learning is proposed. The model exploits Non-Axiomatic Logic (NAL)~\cite{wang2013nal} as the basis of representation, inference, and learning. The model is brain-inspired since the structure is highly inspired by Hierarchical Temporal Memory (HTM)~\cite{hawkins2016seq-mem}, which mimics the mini-columns in neocortex~\cite{buxhoeveden2002minicolumn}, though \textit{mini-column} is imposed of a special meaning (\textit{i.e.}, \textit{concept} in NAL) in this paper. A learning algorithm is designed, comprising of three steps: hypothesizing, revising, and recycling -- Due to the \textit{Assumption of Insufficient Knowledge and Resources} (AIKR)~\cite{wang2019defining}, there is no way to memorize the whole dataset (to do \textit{offline learning}~\cite{hoi2021online}) as well as all possibilities of sequences, in the mean while, the time complexity should be a constant when handling each \textit{event} input to the system. The hypothesizing and recycling procedures are responsible for resources allocation, to guarantee that the model satisfies AIKR. In the revising procedure, candidate \textit{links} are picked out for \textit{temporal induction} and \textit{revision} that are logical rules in NAL. To predict future events, temporal deduction rule is applied to generate anticipations. The model can be converted to \textit{Narsese}, the formal language of NAL, so that the model is fully interpretable, explainable, and even trust-worthy. 

The dataset for test is assumed to be an endless list of events, meaning that theoretically there is no explicit head or tail of the list; thus, the model has to do the so called online learning~\cite{hoi2021online} and work in \textit{real-time} in a sense. The dataset is generated synthetically, with $50\%$ predictable events and $50\%$ random events. For the predictable part, a certain number (denoted as $p$) of prototypes of sequences are generated; each of the prototype has a certain length (denoted as $m$). To test the capacity of the model, with different $m$s and $p$s, the model is asked to make anticipations on next events. The correctness of anticipations is plotted in Fig.~\ref{fig:capacity-tests-simple-case} and Fig.~\ref{fig:capacity-tests}, showing that the model performs well in several settings with different requirements for capacity. In addition, another task, \textit{a.k.a.} continual learning~\cite{zeng2019continual}, is used to test whether the model suffers from \textit{catastrophic forgetting}~\cite{mccloskey1989cf}, which is a long-standing problem in models with distributed representation, such as neural networks. The problem of \textit{catastrophic forgetting} does not occur in the model proposed, as shown in Fig.~\ref{fig:cf-ne=26} and Fig.~\ref{fig:cf-ne=1000}. This is because the model exploits logical representation (or concept-centered representation, see \ref{sec:logic-repr}), through which modifying one \textit{concept} or its relevant connections does not intervene other irrelevant \textit{concepts}, though forgetting is inevitable due to insufficient resources.

This paper demonstrates the potential of learning sequential patterns in a logical way, though there is some interesting work for further researching.

\backmatter
\section*{Declarations}

\bmhead{Supplementary information}

The source code is available at ``\url{https://github.com/bowen-xu/SeL-NAL}''.

\bmhead{Acknowledgments}

I thank those who reviewed this article for their suggestions; especially, I discussed a lot with my advisor, Dr. Pei Wang\footnote{\url{https://cis.temple.edu/~pwang/}}, on the idea and the work proposed in this paper, and I appreciate his comments and advice.

\bibliography{sn-bibliography}


\begin{thebibliography}{26}
\ifx \bisbn   \undefined \def \bisbn  #1{ISBN #1}\fi
\ifx \binits  \undefined \def \binits#1{#1}\fi
\ifx \bauthor  \undefined \def \bauthor#1{#1}\fi
\ifx \batitle  \undefined \def \batitle#1{#1}\fi
\ifx \bjtitle  \undefined \def \bjtitle#1{#1}\fi
\ifx \bvolume  \undefined \def \bvolume#1{\textbf{#1}}\fi
\ifx \byear  \undefined \def \byear#1{#1}\fi
\ifx \bissue  \undefined \def \bissue#1{#1}\fi
\ifx \bfpage  \undefined \def \bfpage#1{#1}\fi
\ifx \blpage  \undefined \def \blpage #1{#1}\fi
\ifx \burl  \undefined \def \burl#1{\textsf{#1}}\fi
\ifx \doiurl  \undefined \def \doiurl#1{\url{https://doi.org/#1}}\fi
\ifx \betal  \undefined \def \betal{\textit{et al.}}\fi
\ifx \binstitute  \undefined \def \binstitute#1{#1}\fi
\ifx \binstitutionaled  \undefined \def \binstitutionaled#1{#1}\fi
\ifx \bctitle  \undefined \def \bctitle#1{#1}\fi
\ifx \beditor  \undefined \def \beditor#1{#1}\fi
\ifx \bpublisher  \undefined \def \bpublisher#1{#1}\fi
\ifx \bbtitle  \undefined \def \bbtitle#1{#1}\fi
\ifx \bedition  \undefined \def \bedition#1{#1}\fi
\ifx \bseriesno  \undefined \def \bseriesno#1{#1}\fi
\ifx \blocation  \undefined \def \blocation#1{#1}\fi
\ifx \bsertitle  \undefined \def \bsertitle#1{#1}\fi
\ifx \bsnm \undefined \def \bsnm#1{#1}\fi
\ifx \bsuffix \undefined \def \bsuffix#1{#1}\fi
\ifx \bparticle \undefined \def \bparticle#1{#1}\fi
\ifx \barticle \undefined \def \barticle#1{#1}\fi
\bibcommenthead
\ifx \bconfdate \undefined \def \bconfdate #1{#1}\fi
\ifx \botherref \undefined \def \botherref #1{#1}\fi
\ifx \url \undefined \def \url#1{\textsf{#1}}\fi
\ifx \bchapter \undefined \def \bchapter#1{#1}\fi
\ifx \bbook \undefined \def \bbook#1{#1}\fi
\ifx \bcomment \undefined \def \bcomment#1{#1}\fi
\ifx \oauthor \undefined \def \oauthor#1{#1}\fi
\ifx \citeauthoryear \undefined \def \citeauthoryear#1{#1}\fi
\ifx \endbibitem  \undefined \def \endbibitem {}\fi
\ifx \bconflocation  \undefined \def \bconflocation#1{#1}\fi
\ifx \arxivurl  \undefined \def \arxivurl#1{\textsf{#1}}\fi
\csname PreBibitemsHook\endcsname

\bibitem[\protect\citeauthoryear{Conway}{2012}]{conway2012seq}
\begin{bchapter}
\bauthor{\bsnm{Conway}, \binits{C.M.}}:
\bctitle{Sequential learning}.
In: \beditor{\bsnm{Seel}, \binits{N.M.}} (ed.)
\bbtitle{Encyclopedia of the Sciences of Learning Sequential Learning},
pp. \bfpage{3047}--\blpage{3050}.
\bpublisher{Springer},
\blocation{LLC, 233 Spring Street, New York, NY 10013, USA}
(\byear{2012})
\end{bchapter}
\endbibitem

\bibitem[\protect\citeauthoryear{Sun and Giles}{2001}]{sun2001sequence}
\begin{barticle}
\bauthor{\bsnm{Sun}, \binits{R.}},
\bauthor{\bsnm{Giles}, \binits{C.L.}}:
\batitle{Sequence learning: From recognition and prediction to sequential
  decision making}.
\bjtitle{IEEE Intelligent Systems}
\bvolume{16}(\bissue{4}),
\bfpage{67}--\blpage{70}
(\byear{2001})
\end{barticle}
\endbibitem

\bibitem[\protect\citeauthoryear{Clegg et~al.}{1998}]{clegg1998sequence}
\begin{barticle}
\bauthor{\bsnm{Clegg}, \binits{B.A.}},
\bauthor{\bsnm{DiGirolamo}, \binits{G.J.}},
\bauthor{\bsnm{Keele}, \binits{S.W.}}:
\batitle{Sequence learning}.
\bjtitle{Trends in Cognitive Sciences}
\bvolume{2}(\bissue{8}),
\bfpage{275}--\blpage{281}
(\byear{1998})
\doiurl{10.1016/S1364-6613(98)01202-9}
\end{barticle}
\endbibitem

\bibitem[\protect\citeauthoryear{Rabiner and Juang}{1986}]{rabiner1986hmm}
\begin{barticle}
\bauthor{\bsnm{Rabiner}, \binits{L.}},
\bauthor{\bsnm{Juang}, \binits{B.}}:
\batitle{An introduction to hidden markov models}.
\bjtitle{IEEE ASSP Magazine}
\bvolume{3}(\bissue{1}),
\bfpage{4}--\blpage{16}
(\byear{1986})
\doiurl{10.1109/MASSP.1986.1165342}
\end{barticle}
\endbibitem

\bibitem[\protect\citeauthoryear{Medsker and Jain}{2001}]{medsker2001rnn}
\begin{barticle}
\bauthor{\bsnm{Medsker}, \binits{L.R.}},
\bauthor{\bsnm{Jain}, \binits{L.}}:
\batitle{Recurrent neural networks}.
\bjtitle{Design and Applications}
\bvolume{5},
\bfpage{64}--\blpage{67}
(\byear{2001})
\end{barticle}
\endbibitem

\bibitem[\protect\citeauthoryear{Vaswani et~al.}{2017}]{vaswani2017transformer}
\begin{botherref}
\oauthor{\bsnm{Vaswani}, \binits{A.}},
\oauthor{\bsnm{Shazeer}, \binits{N.}},
\oauthor{\bsnm{Parmar}, \binits{N.}},
\oauthor{\bsnm{Uszkoreit}, \binits{J.}},
\oauthor{\bsnm{Jones}, \binits{L.}},
\oauthor{\bsnm{Gomez}, \binits{A.N.}},
\oauthor{\bsnm{Kaiser}, \binits{{\L}.}},
\oauthor{\bsnm{Polosukhin}, \binits{I.}}:
Attention is all you need.
Advances in neural information processing systems
\textbf{30}
(2017)
\end{botherref}
\endbibitem

\bibitem[\protect\citeauthoryear{Hawkins and Ahmad}{2016}]{hawkins2016seq-mem}
\begin{botherref}
\oauthor{\bsnm{Hawkins}, \binits{J.}},
\oauthor{\bsnm{Ahmad}, \binits{S.}}:
Why neurons have thousands of synapses, a theory of sequence memory in
  neocortex.
Frontiers in neural circuits,
23
(2016)
\end{botherref}
\endbibitem

\bibitem[\protect\citeauthoryear{Wang}{2013}]{wang2013nal}
\begin{bbook}
\bauthor{\bsnm{Wang}, \binits{P.}}:
\bbtitle{Non-axiomatic Logic: A Model of Intelligent Reasoning}.
\bpublisher{World Scientific},
\blocation{Singapore}
(\byear{2013})
\end{bbook}
\endbibitem

\bibitem[\protect\citeauthoryear{Harnad}{1990}]{harnad1990symbol_grounding}
\begin{barticle}
\bauthor{\bsnm{Harnad}, \binits{S.}}:
\batitle{The symbol grounding problem}.
\bjtitle{Physica D: Nonlinear Phenomena}
\bvolume{42}(\bissue{1-3}),
\bfpage{335}--\blpage{346}
(\byear{1990})
\end{barticle}
\endbibitem

\bibitem[\protect\citeauthoryear{Wang}{2005}]{wang2005semantics}
\begin{barticle}
\bauthor{\bsnm{Wang}, \binits{P.}}:
\batitle{Experience-grounded semantics: a theory for intelligent systems}.
\bjtitle{Cognitive Systems Research}
\bvolume{6}(\bissue{4}),
\bfpage{282}--\blpage{302}
(\byear{2005})
\end{barticle}
\endbibitem

\bibitem[\protect\citeauthoryear{Wang and Hammer}{2015}]{wang2015temporal}
\begin{bchapter}
\bauthor{\bsnm{Wang}, \binits{P.}},
\bauthor{\bsnm{Hammer}, \binits{P.}}:
\bctitle{Issues in temporal and causal inference}.
In: \bbtitle{Artificial General Intelligence: 8th International Conference, AGI
  2015, AGI 2015, Berlin, Germany, July 22-25, 2015, Proceedings 8},
pp. \bfpage{208}--\blpage{217}
(\byear{2015}).
\bcomment{Springer}
\end{bchapter}
\endbibitem

\bibitem[\protect\citeauthoryear{Buxhoeveden and
  Casanova}{2002}]{buxhoeveden2002minicolumn}
\begin{barticle}
\bauthor{\bsnm{Buxhoeveden}, \binits{D.P.}},
\bauthor{\bsnm{Casanova}, \binits{M.F.}}:
\batitle{The minicolumn hypothesis in neuroscience}.
\bjtitle{Brain}
\bvolume{125}(\bissue{5}),
\bfpage{935}--\blpage{951}
(\byear{2002})
\end{barticle}
\endbibitem

\bibitem[\protect\citeauthoryear{Wang}{2001}]{wang2001uncertainty}
\begin{bchapter}
\bauthor{\bsnm{Wang}, \binits{P.}}:
\bctitle{Confidence as higher-order uncertainty}.
In: \bbtitle{Proceedings of the Second International Symposium on Imprecise
  Probabilities and Their Applications},
\bconflocation{Ithaca, New York},
pp. \bfpage{352}--\blpage{361}
(\byear{2001})
\end{bchapter}
\endbibitem

\bibitem[\protect\citeauthoryear{Hoi et~al.}{2021}]{hoi2021online}
\begin{barticle}
\bauthor{\bsnm{Hoi}, \binits{S.C.}},
\bauthor{\bsnm{Sahoo}, \binits{D.}},
\bauthor{\bsnm{Lu}, \binits{J.}},
\bauthor{\bsnm{Zhao}, \binits{P.}}:
\batitle{Online learning: A comprehensive survey}.
\bjtitle{Neurocomputing}
\bvolume{459},
\bfpage{249}--\blpage{289}
(\byear{2021})
\end{barticle}
\endbibitem

\bibitem[\protect\citeauthoryear{McCloskey and Cohen}{1989}]{mccloskey1989cf}
\begin{bchapter}
\bauthor{\bsnm{McCloskey}, \binits{M.}},
\bauthor{\bsnm{Cohen}, \binits{N.J.}}:
\bctitle{Catastrophic {Interference} in {Connectionist} {Networks}: {The}
  {Sequential} {Learning} {Problem}}.
In: \beditor{\bsnm{Bower}, \binits{G.H.}} (ed.)
\bbtitle{Psychology of {Learning} and {Motivation}}
vol. \bseriesno{24},
pp. \bfpage{109}--\blpage{165}.
\bpublisher{Academic Press}, \blocation{???}
(\byear{1989})
\end{bchapter}
\endbibitem

\bibitem[\protect\citeauthoryear{Gerstner et~al.}{2014}]{gerstner2014snn}
\begin{bbook}
\bauthor{\bsnm{Gerstner}, \binits{W.}},
\bauthor{\bsnm{Kistler}, \binits{W.M.}},
\bauthor{\bsnm{Naud}, \binits{R.}},
\bauthor{\bsnm{Paninski}, \binits{L.}}:
\bbtitle{Neuronal Dynamics: From Single Neurons to Networks and Models of
  Cognition}.
\bpublisher{Cambridge University Press},
\blocation{Cambridge}
(\byear{2014})
\end{bbook}
\endbibitem

\bibitem[\protect\citeauthoryear{Hawkins et~al.}{2016}]{hawkins2016bami}
\begin{botherref}
\oauthor{\bsnm{Hawkins}, \binits{J.}},
\oauthor{\bsnm{Ahmad}, \binits{S.}},
\oauthor{\bsnm{Purdy}, \binits{S.}},
\oauthor{\bsnm{Lavin}, \binits{A.}}:
Biological and machine intelligence (bami).
Initial online release 0.4
(2016)
\end{botherref}
\endbibitem

\bibitem[\protect\citeauthoryear{Hammer et~al.}{2016}]{hammer2016opennars}
\begin{bchapter}
\bauthor{\bsnm{Hammer}, \binits{P.}},
\bauthor{\bsnm{Lofthouse}, \binits{T.}},
\bauthor{\bsnm{Wang}, \binits{P.}}:
\bctitle{The opennars implementation of the non-axiomatic reasoning system}.
In: \bbtitle{Artificial General Intelligence: 9th International Conference, AGI
  2016, New York, NY, USA, July 16-19, 2016, Proceedings 9},
pp. \bfpage{160}--\blpage{170}
(\byear{2016}).
\bcomment{Springer}
\end{bchapter}
\endbibitem

\bibitem[\protect\citeauthoryear{Wang}{2020}]{wang2020consciousness}
\begin{barticle}
\bauthor{\bsnm{Wang}, \binits{P.}}:
\batitle{A constructive explanation of consciousness}.
\bjtitle{Journal of Artificial Intelligence and Consciousness}
\bvolume{7}(\bissue{02}),
\bfpage{257}--\blpage{275}
(\byear{2020})
\end{barticle}
\endbibitem

\bibitem[\protect\citeauthoryear{Wang}{2019}]{wang2019defining}
\begin{barticle}
\bauthor{\bsnm{Wang}, \binits{P.}}:
\batitle{On defining artificial intelligence}.
\bjtitle{Journal of Artificial General Intelligence}
\bvolume{10}(\bissue{2}),
\bfpage{1}--\blpage{37}
(\byear{2019})
\end{barticle}
\endbibitem

\bibitem[\protect\citeauthoryear{Zeng et~al.}{2019}]{zeng2019continual}
\begin{barticle}
\bauthor{\bsnm{Zeng}, \binits{G.}},
\bauthor{\bsnm{Chen}, \binits{Y.}},
\bauthor{\bsnm{Cui}, \binits{B.}},
\bauthor{\bsnm{Yu}, \binits{S.}}:
\batitle{Continual learning of context-dependent processing in neural
  networks}.
\bjtitle{Nature Machine Intelligence}
\bvolume{1}(\bissue{8}),
\bfpage{364}--\blpage{372}
(\byear{2019})
\end{barticle}
\endbibitem

\bibitem[\protect\citeauthoryear{Riesenhuber and
  Poggio}{1999}]{riesenhuber1999obj-recog}
\begin{barticle}
\bauthor{\bsnm{Riesenhuber}, \binits{M.}},
\bauthor{\bsnm{Poggio}, \binits{T.}}:
\batitle{Hierarchical models of object recognition in cortex}.
\bjtitle{Nature Neuroscience}
\bvolume{2}(\bissue{11}),
\bfpage{1019}--\blpage{1025}
(\byear{1999})
\end{barticle}
\endbibitem

\bibitem[\protect\citeauthoryear{Lee et~al.}{1999}]{lee1999wta}
\begin{barticle}
\bauthor{\bsnm{Lee}, \binits{D.K.}},
\bauthor{\bsnm{Itti}, \binits{L.}},
\bauthor{\bsnm{Koch}, \binits{C.}},
\bauthor{\bsnm{Braun}, \binits{J.}}:
\batitle{Attention activates winner-take-all competition among visual filters}.
\bjtitle{Nature Neuroscience}
\bvolume{2}(\bissue{4}),
\bfpage{375}--\blpage{381}
(\byear{1999})
\end{barticle}
\endbibitem

\bibitem[\protect\citeauthoryear{Müller et~al.}{2015}]{muller2015piaget}
\begin{bchapter}
\bauthor{\bsnm{Müller}, \binits{U.}},
\bauthor{\bsnm{Ten~Eycke}, \binits{K.}},
\bauthor{\bsnm{Baker}, \binits{L.}}:
\bctitle{Piaget’s {Theory} of {Intelligence}}.
In: \beditor{\bsnm{Goldstein}, \binits{S.}},
\beditor{\bsnm{Princiotta}, \binits{D.}},
\beditor{\bsnm{Naglieri}, \binits{J.A.}} (eds.)
\bbtitle{Handbook of {Intelligence}: {Evolutionary} {Theory}, {Historical}
  {Perspective}, and {Current} {Concepts}},
pp. \bfpage{137}--\blpage{151}.
\bpublisher{Springer},
\blocation{New York, NY}
(\byear{2015})
\end{bchapter}
\endbibitem

\bibitem[\protect\citeauthoryear{Lashley}{1951}]{lashley1951hierachy}
\begin{bchapter}
\bauthor{\bsnm{Lashley}, \binits{K.S.}}:
\bctitle{The problem of serial order in behavior}.
In: \bbtitle{Cerebral Mechanisms in Behavior; the {Hixon} {Symposium}},
pp. \bfpage{112}--\blpage{146}.
\bpublisher{Wiley},
\blocation{Oxford, England}
(\byear{1951})
\end{bchapter}
\endbibitem

\bibitem[\protect\citeauthoryear{Liang et~al.}{2020}]{liang2020sl-snn}
\begin{barticle}
\bauthor{\bsnm{Liang}, \binits{Q.}},
\bauthor{\bsnm{Zeng}, \binits{Y.}},
\bauthor{\bsnm{Xu}, \binits{B.}}:
\batitle{Temporal-sequential learning with a brain-inspired spiking neural
  network and its application to musical memory}.
\bjtitle{Frontiers in Computational Neuroscience}
\bvolume{14},
\bfpage{51}
(\byear{2020})
\end{barticle}
\endbibitem

\end{thebibliography}

\end{document}